\documentclass{article}
\usepackage{times}
\usepackage[margin=1in]{geometry}
\usepackage[round,authoryear]{natbib}
\usepackage{amsmath,amsfonts,bm}

\def\eqref#1{equation~\ref{#1}}
\def\1{\bm{1}}

\DeclareMathAlphabet{\mathsfit}{\encodingdefault}{\sfdefault}{m}{sl}
\SetMathAlphabet{\mathsfit}{bold}{\encodingdefault}{\sfdefault}{bx}{n}

\newcommand{\E}{\mathbb{E}}

\usepackage{amsmath,amssymb,mathtools}
\usepackage{booktabs,multirow,array}
\usepackage{graphicx}
\usepackage{xcolor}
\usepackage{colortbl}
\usepackage{microtype}
\usepackage{enumitem}
\usepackage{needspace}
\usepackage{placeins}
\usepackage{float}
\usepackage{hyperref}
\usepackage{url}
\hypersetup{hidelinks}

\title{Simulating Respondents, Not Single Questions:\\
Coherent Survey Generation with Large Language Models}

\author{Ji Huang\\School of Computer Science, University of Science and Technology of China
\and Mengfei Li\\School of Management, Fudan University
\and Shuai Shao\\School of Computer Science, University of Science and Technology of China}
\date{September 2026}

\newcommand{\method}{\textsc{FR-LLM}}
\newcommand{\mcjp}{\textsc{MCJP}}
\newcommand{\singleft}{\mbox{Single FT}}
\newcommand{\sequentialft}{\mbox{Sequential FT}}
\definecolor{frgray}{gray}{0.93}
\newcommand{\jsd}{\operatorname{JSD}}
\newcommand{\kl}{D_{\mathrm{KL}}}

\begin{document}
\maketitle

\begin{abstract}
Large language models are increasingly used to simulate response distributions in social surveys. Prior work has achieved accurate population-level simulation for individual questions. Real-world questionnaires, however, typically require each respondent to answer a sequence of related questions. A simulated respondent should therefore exhibit coherent preferences across the entire questionnaire, rather than merely produce accurate distributions for isolated items. Existing single-item simulation methods can closely match item-level response distributions, but they do not accurately reproduce how the same respondent answers a complete survey. To address this limitation, we propose FullRespondent-LLM (FR-LLM), a framework for simulating complete virtual survey respondents. FR-LLM first fine-tunes two specialized LLMs: a marginal model that estimates the response distribution of each item and a respondent-level autoregressive model that captures dependencies among answers across the questionnaire. We combine these models through Marginal-Constrained Joint Projection (MCJP), which projects the autoregressive joint distribution onto the set of distributions satisfying the item-level marginals learned by the marginal model. By modeling item distributions and cross-item relationships separately, FR-LLM generates complete questionnaires that reproduce realistic cross-item relationships while retaining the item-level accuracy of strong single-item simulators. Experiments on two real-world social survey datasets show that FR-LLM more accurately reproduces response patterns across multiple questions, while maintaining competitive single-item accuracy and generalizing better to unseen respondent populations and survey questions. We also analyze a small commercial-survey dataset, use the responses generated by each method to make the same pricing and stocking decision, and compare the resulting profits. FR-LLM produces the highest profit, demonstrating its potential for practical survey-based decision making.
\end{abstract}

\section{Introduction}

Large language models (LLMs) are increasingly used to simulate how people answer surveys. Moving beyond prompting, Cao et al.\ \citeyearpar{cao2025specializing} fine-tune first-token probabilities of valid options against country-level answer distributions; Suh et al.\ \citeyearpar{suh2025scaled} train on subpopulation--question response distributions. Huang et al.\ \citeyearpar{huang2026distribution} further align how answer distributions change across respondent backgrounds, improving single-question simulation.

\begin{figure}[t]
    \centering
    \includegraphics[width=0.93\textwidth]{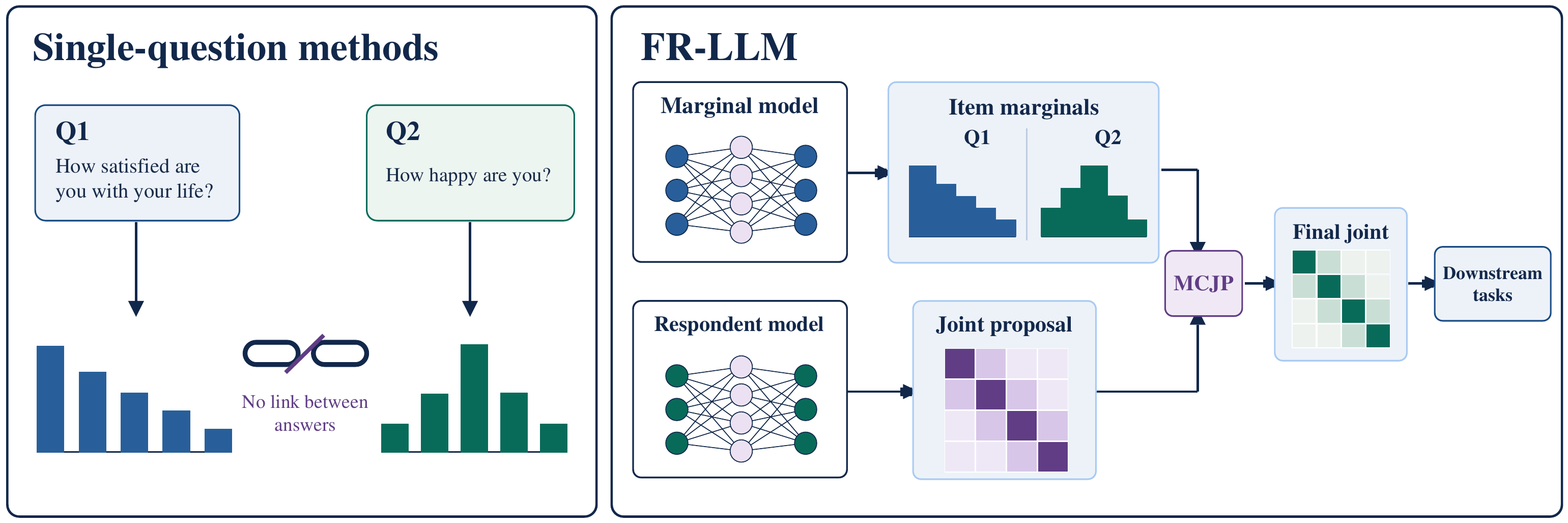}
    \caption{A conceptual satisfaction--happiness example (not an ESS or TALIS item pair). Matching each item's response rates does not ensure that the same simulated respondent gives coherent answers to both; \method{} models and calibrates complete response vectors.}
    \label{fig:motivation}
\end{figure}

Given a respondent's background and one question, these methods can accurately estimate how a population answers that item. A real survey, however, asks the \emph{same person} several related questions, and its applications often depend on a plausible complete set of answers. Someone reporting high life satisfaction, for example, will often also report happiness. A simulator that predicts the two questions separately may match the proportion of satisfied and happy people while failing to reproduce which answers come from the same person (Figure~\ref{fig:motivation}). Such questionnaires can look accurate one item at a time but be misleading when used to group respondents or inform decisions based on multiple answers. They can also have the wrong reliability: Cronbach's $\alpha$ depends on how answers to different questions vary together, which their separate response distributions cannot reveal \citep{williams2026beyond,lukauskas2026plausible}.

\Needspace{7\baselineskip}
A natural alternative is to answer the questions one after another, showing the model its earlier answers before it answers the next question. This lets it learn how a person's answers relate. But learning to produce a complete questionnaire does not separately ensure that the overall proportion choosing each option for each question matches the real survey. In our experiments, this fixed-order sequential approach better captures relationships between answers than single-question fine-tuning, yet usually matches each question's response distribution less accurately.

\Needspace{6\baselineskip}
We propose \method{} to learn these two aspects separately and then bring them together: one model learns how often each answer is chosen, while another learns which answers tend to occur in the same person. The first LLM is fine-tuned on individual background--question--answer observations and estimates each question's response distribution. The second is fine-tuned on complete questionnaires from real respondents; it learns to answer using previous responses as context, with question order randomized during training, and generates complete candidate questionnaires. At inference, Marginal-Constrained Joint Projection (\mcjp{}) adjusts how often each candidate is selected so that the final population matches the first model's item-level predictions. Formally, it chooses the closest distribution to the respondent model's proposal in reverse KL. Because it reweights and resamples \emph{whole} questionnaires rather than changing individual answers, the final virtual respondents retain the relationships learned by the second model while matching the single-question distributions learned by the first.

We evaluate \method{} on ESS11 and TALIS 2018 with two LLM families under held-out questions, respondent populations, and both together. It improves multi-item accuracy while maintaining competitive item-level accuracy. On a separately collected commercial-survey dataset, we use synthetic responses to choose a selling price and stocking quantity, then assess profit on held-out responses. \method{} produces the highest profit. Our contributions are:
\begin{itemize}[leftmargin=*,topsep=2pt,itemsep=1pt]
    \item We introduce \method{}, which combines separately learned item marginals and complete-response proposals through constrained projection for questionnaire-level tasks.
    \item We evaluate questionnaire reliability and validity with Cronbach's $\alpha$ and AVE alongside item and joint distributional errors.
    \item Across two large social surveys and a small commercial decision case, we demonstrate stronger multi-item accuracy, competitive item accuracy, and the practical value of coherent virtual respondents.
\end{itemize}

\section{FullRespondent-LLM}
\label{sec:method}

\subsection{Background and notation}

A \emph{construct} is an attribute measured by several related items, such as self-efficacy across classroom situations. Survey-defined item groupings are used for evaluation; their names and definitions are withheld from the models.

For $r$ scored items, Cronbach's $\alpha=\frac{r}{r-1}\left(1-\frac{\sum_j\operatorname{Var}(Y_j)}{\operatorname{Var}(\sum_jY_j)}\right)$ measures internal consistency \citep{cronbach1951coefficient}. Average variance extracted, $\mathrm{AVE}=r^{-1}\sum_j\lambda_j^2$ for standardized one-factor loadings $\lambda_j$, summarizes convergent validity \citep{fornell1981evaluating}. Both depend on linked answers. We compare their generated and human values across constructs as diagnostics, not training targets or stand-alone proof of validity.

\begin{figure}[t]
    \centering
    \includegraphics[width=0.99\textwidth]{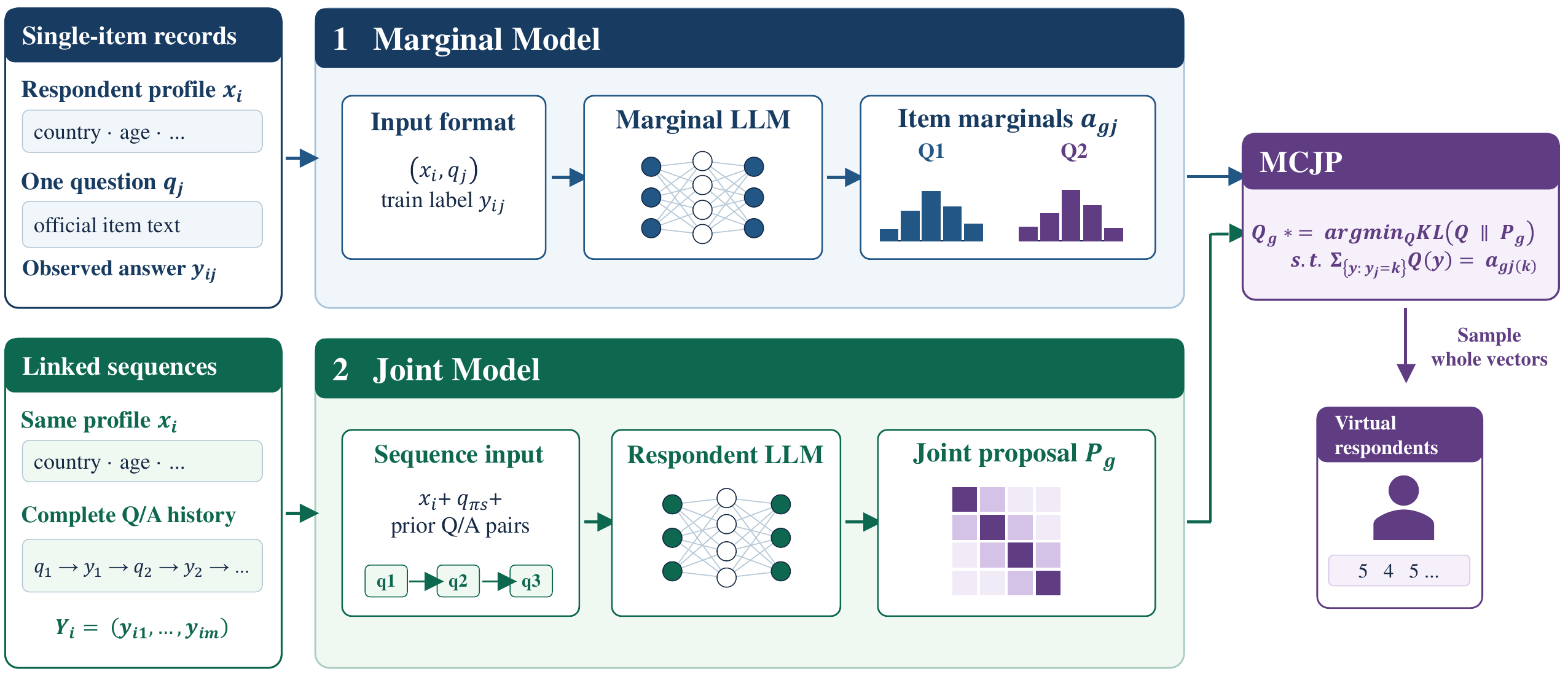}
    \caption{\method{} training and generation. Stage 1 scores one observed item at a time to estimate population marginals. Stage 2 trains on linked question--answer sequences from each respondent to propose complete vectors. \mcjp{} reweights those vectors to match the learned marginals within each populated cell; it does not replace individual answers.}
    \label{fig:method}
\end{figure}

Let $q_j$ and $Y_j\in\{1,\ldots,K_j\}$ be item $j$ and its answer, and $Y=(Y_1,\ldots,Y_m)$ a complete response. Respondent $i$ has background $x_i$ and linked answers $Y_i$; $g=G(x)$ is a prespecified population cell. We target the human joint $T_g(y)=p_{\mathrm{human}}(Y=y\mid G(X)=g,q_{1:m})$, not just item marginals. At test time we receive backgrounds but no target answers or construct names. Figure~\ref{fig:method} shows how the stages learn margins, propose complete responses, and combine them; only the second LLM sees prior answers.

\subsection{Stage 1: learning item marginals}

This stage aims to reproduce how often each option is chosen in a population cell. Accurate item rates matter even when the eventual output is a complete questionnaire, but a model trained on isolated questions cannot learn which answers tend to occur together. We therefore train a separate LoRA-adapted LLM \citep{hu2022lora} on one observed item at a time. Given respondent background $x_i$, item $q_j$, and its response scale (Appendix Figure~\ref{fig:prompt_single}), it scores only valid response tokens to produce probabilities $a_{ij}(k)=\operatorname{softmax}_k f_\phi(x_i,q_j)$. Ordinary cross-entropy minimizes $\mathcal L_{\mathrm{marg}}(\phi)=\mathbb E[-\log a_{ij}(Y_{ij})]$ over observed respondent--item pairs. In expectation, this is the entropy of the human item distribution plus its forward KL to the model distribution; hard answers thus provide samples for learning population rates.

At inference, the model predicts each item for the frozen target backgrounds. Within cell $g$, we average these probabilities using normalized protocol weights $\bar w_i$, obtaining the anchor $a_{gj}(k)=\sum_{i:G(x_i)=g}\bar w_i a_{ij}(k)$. This averaging retains the target cell's background composition without using any held-out answers. The anchors specify desired item rates for Stage 3, not respondent-level answer vectors.

\begin{figure}[t]
    \centering
    \includegraphics[width=0.6\linewidth]{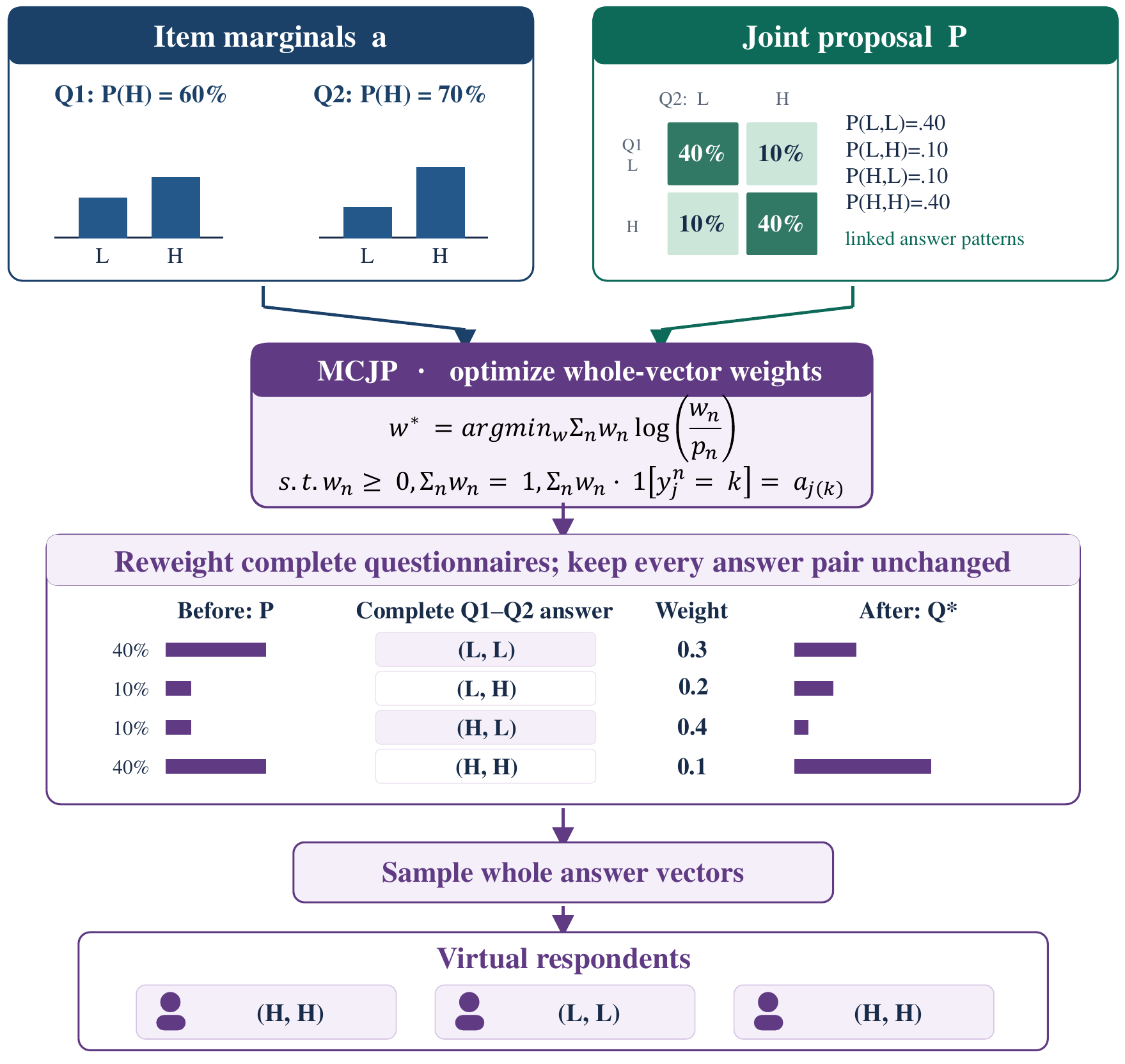}
    \caption{Schematic of \mcjp{} on two binary questions, not an ESS or TALIS evaluation. Stage 1 supplies item targets, the respondent model supplies linked answer pairs, and projection reweights complete pairs before resampling. The displayed weight column is illustrative, not the numerical optimum for the displayed targets; actual weights are determined by Eq.~\ref{eq:iproj} and IPF.}
    \label{fig:mcjp}
\end{figure}

\subsection{Stage 2: learning complete respondents}

This stage provides what isolated-item training cannot: linked answers that reflect a single respondent's preferences across questions. The respondent model uses the same background and item wording, but also sees that person's preceding question--answer pairs (Appendix Figure~\ref{fig:prompt_joint}). In a sampled item order $\pi$, it predicts each answer conditional on the previous answers, so the product of these conditional probabilities defines a joint distribution $P_\theta^\pi$ over complete questionnaires. Training teacher-forces the observed history, sums the losses over all items for one respondent, and performs one backward pass:
\begin{equation}
    \mathcal L_{\mathrm{joint}}(\theta)=
    \mathbb E_{(x_i,Y_i),\pi}\left[-\sum_{s=1}^{m}
    \log p_\theta(Y_{\pi_s}\mid x_i,q_{\pi_{\le s}},Y_{\pi_{<s}})\right].
    \label{eq:jointloss}
\end{equation}
By the probability chain rule, this is a complete-response likelihood objective, averaged over sampled orders rather than tied to the questionnaire display order. At inference, the model uses its own generated histories and a fresh order for each candidate. Averaging across orders and the target backgrounds gives the cell-level proposal
\begin{equation}
    P_g(y)=\sum_{i:G(x_i)=g}\bar w_i\,
    \mathbb E_{\pi\sim\Pi}P_\theta^\pi(y\mid x_i,q_{1:m}).
    \label{eq:mixtureproposal}
\end{equation}
This proposal retains linked answer patterns. It need not, however, reproduce each item's response rate as accurately as Stage 1, which motivates the final combination.

\subsection{Stage 3: Marginal-Constrained Joint Projection}

The two trained models offer complementary information: Stage 1 supplies accurate item rates, while Stage 2 supplies plausible combinations of answers. This stage combines them without retraining either model or independently replacing answers. For a populated cell, let $P=P_g$ be the proposal in Eq.~\ref{eq:mixtureproposal} and $a=(a_{gj})_j$ its Stage-1 target rates. Write $\varphi_{jk}(y)=\mathbb 1\{y_j=k\}$ for each item and all but one redundant option, and let $\mathcal C(a)$ contain joint distributions whose expected indicators equal $a$. \mcjp{} finds the distribution closest to the proposal subject to those rates:
\begin{equation}
    Q_a=\arg\min_{Q\in\mathcal C(a)}\kl(Q\|P).
    \label{eq:iproj}
\end{equation}
Here reverse KL means changing the proposal only as needed to satisfy the calibrated rates. With feasible interior margins, the solution has the form $Q_a(y)\propto P(y)\exp\{\lambda_a^\top\varphi(y)\}$, with weights chosen so its item marginals equal $a$. These item--option weights reweight whole answer vectors; they do not fit another model of cross-item interactions.

Figure~\ref{fig:mcjp} illustrates the reweighting of entire answer patterns. In practice, we sample $M=16$ complete candidates per respondent with fresh item orders and pool them within populated cells. Iterative proportional fitting (IPF) adjusts their weights to match $a$, after which randomized systematic resampling produces three rollouts of whole vectors. If the candidate pool lacks the support needed for exact matching, we report an adjustment to the nearest feasible margin. No held-out human response enters \mcjp{}.

\paragraph{Projection guarantee.}
The projection also separates two sources of error. Let $T$ be the human joint, $t=\mathbb E_T\varphi$ its true margins, and $Q_t$ the oracle projection onto $\mathcal C(t)$. Under full support and feasible interior margins,
\begin{equation}
    \boxed{\kl(T\|Q_a)=
    \underbrace{\kl(T\|Q_t)}_{\text{dependence residual}}+
    \underbrace{\kl(Q_t\|Q_a)}_{\text{marginal-anchor error}}.}
    \label{eq:decomp}
\end{equation}
Thus the final discrepancy consists of a residual dependence error and an error due to imperfect marginal anchors. Exact anchors cannot worsen forward KL relative to the proposal; with imperfect anchors, the outcome depends on their accuracy. The ideal reweighting also preserves conditional odds ratios. Appendix~\ref{app:theory} gives the proof, a quantitative bound for anchor error, and the finite-support qualification relevant to IPF.

\FloatBarrier
\section{Experiments}
\label{sec:setup}

\subsection{Settings}

\paragraph{Datasets.}
We evaluate on ESS11's 21-item Human Values Scale \citep{ess11data,schwartz2012refining} and TALIS 2018's 12-item teacher self-efficacy battery \citep{oecd2019talis}. ESS11 trains on four constructs (9 items) and holds out six (12 items); TALIS trains on two constructs (8 items) and holds out one (4 items). \textbf{M1} tests unseen constructs in seen populations, \textbf{M2} tests seen constructs in unseen countries, and \textbf{M3} tests both unseen constructs and countries. Table~\ref{tab:samplecounts} gives the respondent counts. Development countries are disjoint from test countries and used only for checkpoint selection; methods share the same target respondents within each mode. We additionally use 488 records from our Little-Treat Consumption survey for a separate held-out pricing-and-stocking task, not the ESS11/TALIS transfer benchmark (Appendix~\ref{app:littletreat}).

\begin{table}[t]
    \caption{Distinct respondents by split. M1: unseen constructs; M2: unseen countries; M3: both. M2 and M3 share respondents but use different items.}
    \label{tab:samplecounts}
    \centering
    \small
    \begin{tabular*}{\textwidth}{@{\extracolsep{\fill}}lrrrr@{}}
        \toprule
        Dataset & Train $n$ & M1 test $n$ & M2 test $n$ & M3 test $n$ \\
        \midrule
        ESS11 & 28,268 & 28,268 & 7,312 & 7,312 \\
        TALIS 2018 & 12,000 & 8,000 & 8,000 & 8,000 \\
        \bottomrule
    \end{tabular*}
\end{table}

\paragraph{Models.}
We use Qwen3.5-9B \citep{qwen35model} and Ministral-3-8B \citep{ministralmodel}. Learned methods share construct-free item wording and valid-option scoring. LoRA settings, prompts, and optimization are in Appendix~\ref{app:implementation}; checkpoints use development countries, and reported results average three generation rollouts.

\Needspace{8\baselineskip}
\paragraph{Metrics.}
All four errors are lower-is-better. Item JSD compares human and generated answer distributions per question; Construct JSD compares complete answer tuples within each construct, testing cross-item patterns. $\alpha$ and one-factor AVE MAE average the absolute human--generated difference across constructs, measuring agreement in reliability and convergent-validity summaries. JSD is calculated before averaging; Appendix~\ref{app:metrics} specifies weights and aggregation.

\paragraph{Baselines.}
\textbf{Zero-shot} scores valid answers without survey fine-tuning. \textbf{\singleft{}} follows the single-question setting of \citet{cao2025specializing,suh2025scaled}: it fine-tunes and samples each item independently, without answer history. \textbf{\sequentialft{}} extends their question-wise fine-tuning setting to complete questionnaires: it processes items in a fixed order, conditions on preceding answers, and makes a separate update for each item. This fixed-order, history-conditioned variant is our baseline implementation, not a method claimed by those papers.

\subsection{Main results}
\label{sec:results}

\begin{table}[t]
\caption{ESS11 results for both backbone models. Lower is better. Bold is best and underline is second best within each backbone and transfer mode; ties receive the same mark.}
\label{tab:essmain}
\centering
\footnotesize
\setlength{\tabcolsep}{3.8pt}
\renewcommand{\arraystretch}{1.16}
\begin{tabular*}{\textwidth}{@{\extracolsep{\fill}}>{\centering\arraybackslash}m{0.055\textwidth}>{\raggedright\arraybackslash}m{0.16\textwidth}*{2}{>{\centering\arraybackslash}m{0.10\textwidth}>{\centering\arraybackslash}m{0.05\textwidth}>{\centering\arraybackslash}m{0.05\textwidth}>{\centering\arraybackslash}m{0.09\textwidth}}@{}}
\toprule
\multirow[c]{2}{*}{Mode} & \multirow[c]{2}{*}{Method} & \multicolumn{4}{c}{Qwen3.5-9B} & \multicolumn{4}{c}{Ministral-3-8B} \\
\cmidrule(lr){3-6}\cmidrule(lr){7-10}
 & & Cronbach's $\alpha$ MAE & AVE MAE & Item JSD & Construct JSD & Cronbach's $\alpha$ MAE & AVE MAE & Item JSD & Construct JSD \\
\midrule
\multirow{4}{*}{M1} & Zero-shot & .507 & .348 & .075 & .124 & .442 & .316 & .106 & .168 \\
 & \mbox{Single FT} & .363 & .272 & \textbf{.024} & \underline{.037} & .367 & .274 & \textbf{.031} & \underline{.048} \\
 & \mbox{Sequential FT} & \underline{.084} & \underline{.075} & \underline{.044} & .047 & \underline{.114} & \underline{.103} & \underline{.089} & .091 \\
\rowcolor{frgray}  & \method{} & \textbf{.078} & \textbf{.071} & \textbf{.024} & \textbf{.021} & \textbf{.046} & \textbf{.040} & \textbf{.031} & \textbf{.032} \\
\midrule
\multirow{4}{*}{M2} & Zero-shot & .625 & .405 & .084 & .175 & .524 & .370 & .107 & .201 \\
 & \mbox{Single FT} & .493 & .346 & \textbf{.014} & \underline{.039} & .496 & .346 & \textbf{.013} & \underline{.041} \\
 & \mbox{Sequential FT} & \underline{.089} & \underline{.082} & \underline{.045} & .049 & \underline{.111} & \underline{.103} & \underline{.052} & .048 \\
\rowcolor{frgray}  & \method{} & \textbf{.028} & \textbf{.030} & \textbf{.014} & \textbf{.009} & \textbf{.044} & \textbf{.050} & \textbf{.013} & \textbf{.011} \\
\midrule
\multirow{4}{*}{M3} & Zero-shot & .470 & .319 & .078 & .141 & .399 & .284 & .110 & .183 \\
 & \mbox{Single FT} & .378 & .273 & \textbf{.038} & \underline{.061} & .375 & .271 & \textbf{.046} & \underline{.072} \\
 & \mbox{Sequential FT} & \underline{.099} & \underline{.084} & \underline{.054} & .075 & \underline{.130} & \underline{.113} & \underline{.090} & .102 \\
\rowcolor{frgray}  & \method{} & \textbf{.078} & \textbf{.066} & \textbf{.038} & \textbf{.045} & \textbf{.047} & \textbf{.037} & \textbf{.046} & \textbf{.056} \\
\bottomrule
\end{tabular*}
\end{table}

\begin{table}[t]
\caption{TALIS 2018 results for both backbone models. Lower is better. Bold is best and underline is second best within each backbone and transfer mode; ties receive the same mark.}
\label{tab:talismain}
\centering
\footnotesize
\setlength{\tabcolsep}{3.5pt}
\renewcommand{\arraystretch}{1.16}
\begin{tabular*}{\textwidth}{@{\extracolsep{\fill}}>{\centering\arraybackslash}m{0.055\textwidth}>{\raggedright\arraybackslash}m{0.16\textwidth}*{2}{>{\centering\arraybackslash}m{0.10\textwidth}>{\centering\arraybackslash}m{0.06\textwidth}>{\centering\arraybackslash}m{0.06\textwidth}>{\centering\arraybackslash}m{0.09\textwidth}}@{}}
\toprule
\multirow[c]{2}{*}{Mode} & \multirow[c]{2}{*}{Method} & \multicolumn{4}{c}{Qwen3.5-9B} & \multicolumn{4}{c}{Ministral-3-8B} \\
\cmidrule(lr){3-6}\cmidrule(lr){7-10}
 & & Cronbach's $\alpha$ MAE & AVE MAE & Item JSD & Construct JSD & Cronbach's $\alpha$ MAE & AVE MAE & Item JSD & Construct JSD \\
\midrule
\multirow{4}{*}{M1} & Zero-shot & .809 & .395 & .061 & .296 & .808 & .392 & .141 & .474 \\
 & \mbox{Single FT} & .642 & .362 & \underline{.017} & .144 & .831 & .396 & \textbf{.020} & .168 \\
 & \mbox{Sequential FT} & \underline{.094} & \underline{.142} & \textbf{.015} & \underline{.058} & \underline{.114} & \underline{.181} & \underline{.029} & \underline{.075} \\
\rowcolor{frgray}  & \method{} & \textbf{.008} & \textbf{.007} & \underline{.017} & \textbf{.043} & \textbf{.049} & \textbf{.058} & \textbf{.020} & \textbf{.052} \\
\midrule
\multirow{4}{*}{M2} & Zero-shot & .786 & .387 & .069 & .309 & .811 & .388 & .150 & .486 \\
 & \mbox{Single FT} & .772 & .383 & \textbf{.012} & .150 & .817 & .387 & \textbf{.016} & .157 \\
 & \mbox{Sequential FT} & \underline{.076} & \underline{.110} & \underline{.015} & \underline{.034} & \underline{.114} & \underline{.180} & \underline{.026} & \underline{.069} \\
\rowcolor{frgray}  & \method{} & \textbf{.003} & \textbf{.004} & \textbf{.012} & \textbf{.013} & \textbf{.037} & \textbf{.042} & \textbf{.016} & \textbf{.015} \\
\midrule
\multirow{4}{*}{M3} & Zero-shot & .796 & .392 & .078 & .311 & .821 & .393 & .149 & .486 \\
 & \mbox{Single FT} & .776 & .388 & \textbf{.018} & .159 & .833 & .393 & \textbf{.023} & .174 \\
 & \mbox{Sequential FT} & \underline{.095} & \underline{.141} & \underline{.023} & \underline{.040} & \underline{.118} & \underline{.186} & \underline{.036} & \underline{.062} \\
\rowcolor{frgray}  & \method{} & \textbf{.001} & \textbf{.003} & \textbf{.018} & \textbf{.029} & \textbf{.045} & \textbf{.055} & \textbf{.023} & \textbf{.046} \\
\bottomrule
\end{tabular*}
\end{table}

Tables~\ref{tab:essmain} and~\ref{tab:talismain} compare both backbones in all three modes on ESS11 and TALIS. Across the resulting 12 settings, \method{} has the lowest Cronbach's $\alpha$ MAE, AVE MAE, and Construct JSD in every case. Its unrounded Item JSD is best or tied in 7 settings, and best or tied in 11 at the tables' three-decimal precision. Thus the gain in complete-response accuracy usually comes with competitive single-question accuracy. The clearest item-level exception is TALIS M1 with Qwen: \sequentialft{} attains Item JSD .015 versus .017 for \method{}, but \method{} improves Construct JSD from .058 to .043 and $\alpha$/AVE MAE from .094/.142 to .008/.007.

The baselines expose the trade-off. \singleft{} matches item distributions but has large $\alpha$ and AVE errors, so independent answers poorly represent complete respondents. \sequentialft{} improves these errors by conditioning on earlier answers, but worsens Item JSD relative to \singleft{} in 11 of 12 settings. \method{} instead preserves competitive item accuracy while improving construct-level joint responses across both surveys.

\subsection{Low-data results}

Many real surveys collect relatively few complete questionnaires. We therefore repeat ESS11 M1 with Qwen using 1,000 training respondents and evaluate the low-data methods on the same 1,000 target respondents. As Table~\ref{tab:lowdata} shows, \method{} retains \singleft{}'s item-level accuracy while reproducing the within-construct joint distribution and questionnaire-level reliability and validity summaries more accurately. Thus, even a small linked sample can provide useful information about which answers occur together, making complete-respondent simulation plausible for narrower survey applications. The full-data rows reproduce Table~\ref{tab:essmain} only as context: they use a larger test population and three rollouts rather than two, so the cross-row differences are not a controlled learning curve.

\begin{table}[t]
    \caption{ESS11 M1 low-data results (Qwen). Full-data rows reproduce Table~\ref{tab:essmain}; bold marks the best low-data value. Lower is better.}
    \label{tab:lowdata}
    \centering\footnotesize
    \setlength{\tabcolsep}{5pt}
    \begin{tabular*}{\textwidth}{@{\extracolsep{\fill}}lrrrrrr@{}}
        \toprule
        Method & Train $n$ & Test $n$ & $\alpha$ MAE & AVE MAE & Item JSD & Construct JSD \\
        \midrule
        Zero-shot & 0 & 1,000 & .494 & .329 & .073 & .131 \\
        \singleft{} & 1,000 & 1,000 & .406 & .290 & \textbf{.054} & .042 \\
        \method{} & 1,000 & 1,000 & \textbf{.066} & \textbf{.061} & \textbf{.054} & \textbf{.022} \\
        \midrule
        \singleft{} & 28,268 & 28,268 & .363 & .272 & .024 & .037 \\
        \method{} & 28,268 & 28,268 & .078 & .071 & .024 & .021 \\
        \bottomrule
    \end{tabular*}
\end{table}

\subsection{A commercial pricing and stocking simulation}

\begin{figure}[t]
    \centering
    \includegraphics[width=\textwidth]{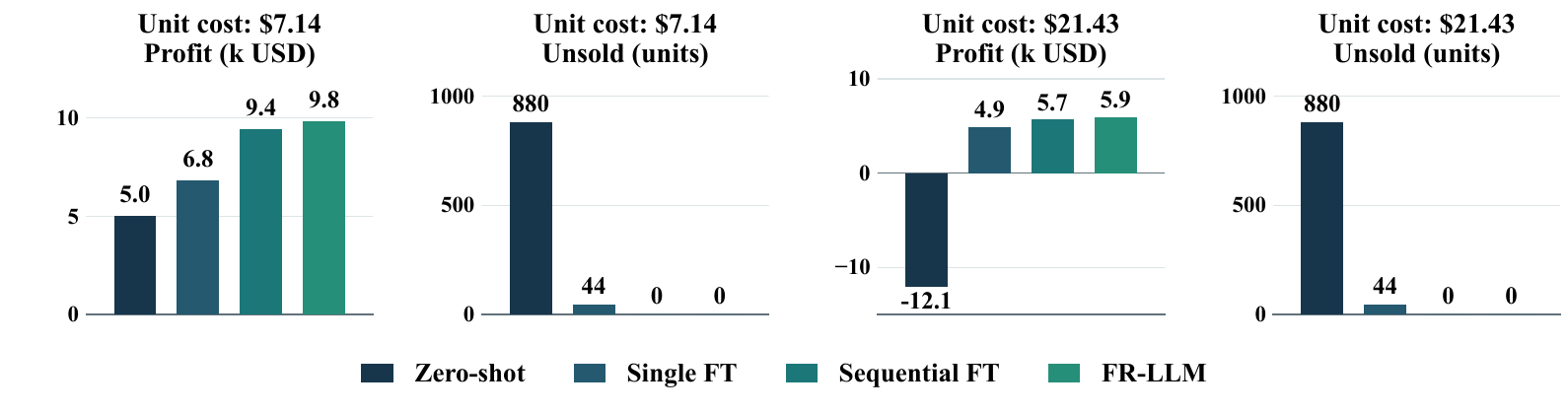}
    \caption{Little-Treat pricing and stocking: profit (thousand USD) and unsold units per 1,000 prospective customers at two unit costs; zero salvage.}
    \label{fig:commercial}
\end{figure}

Beyond matching survey distributions, we test whether synthetic questionnaires support a downstream pricing-and-stocking decision. The Little-Treat survey has 294 training, 97 development, and 97 held-out test respondents. Each model generates six attitude answers, purchase frequency, and the price range of the respondent's most recent purchase. We use frequency to estimate monthly demand and the historical price range as a \emph{proxy} spending ceiling, not an elicited willingness to pay. At each candidate price $p$, the model estimates demand $\widehat D_m(p)$ from these two linked answers, chooses the price maximizing $(p-c)\widehat D_m(p)$ at unit cost $c$, and stocks $\widehat D_m(p)$ units per prospective customer. Applying that decision to held-out human demand gives sales, unsold inventory, and profit after stocking costs. Figure~\ref{fig:commercial} reports two post-hoc cost scenarios with zero salvage, extrapolated to 1,000 customers. Monetary values are USD equivalents at a fixed 7 CNY per USD; the original-currency optimization and quantities are unchanged.

\method{} has the highest profit point estimate at both costs and avoids overstocking, whereas Zero-shot overestimates demand and leaves about 880 units unsold per 1,000 prospective customers. The advantage over \sequentialft{} is small: paired bootstrap intervals over the 97 held-out respondents include zero. This is a conditional decision simulation using historical spending, not evidence of causal price elasticity, realized market profit, or a statistically resolved advantage over that baseline.

\subsection{Analysis}

\textbf{Background shifts.} We change education, age, residence, or two background fields in otherwise fixed ESS11 profiles and compare the resulting item-mean changes with the corresponding human group differences. Shift MAE averages the absolute errors across held-out questions, capturing both the direction and size of each change. Table~\ref{tab:shiftmae} shows that \method{} has the lowest macro-average error across the five contrasts shared by all three methods (.164, versus .169 for \singleft{} and .262 for Zero-shot). It leads on four contrasts, but \singleft{} is better on the combined residence--gender change. These prompt edits test conditional associations, not causal demographic effects; the excluded gender-only contrast is discussed in Appendix~\ref{app:backgroundshift}.

\begin{table}[t]
    \centering
    \caption{ESS11 background-shift MAE across five contrasts (response-scale points; lower is better).}
    \label{tab:shiftmae}
    \footnotesize
    \begin{tabular}{lrrr}
        \toprule
        Background change & Zero-shot & \singleft{} & \method{} \\
        \midrule
        Low to high education & .261 & \underline{.153} & \textbf{.149} \\
        Age 18--29 to 60+ & .399 & \underline{.231} & \textbf{.216} \\
        Urban to rural & .123 & \underline{.112} & \textbf{.111} \\
        Low/young to high/old & .298 & \underline{.198} & \textbf{.177} \\
        Urban/male to rural/female & .229 & \textbf{.150} & \underline{.168} \\
        \midrule
        Macro average & .262 & \underline{.169} & \textbf{.164} \\
        \bottomrule
    \end{tabular}
\end{table}

\textbf{Human sampling reference.} We estimate ordinary sampling variation from 1,000 pairs of independent, stratified respondent-level ESS11 M1 resamples. Table~\ref{tab:humanref} reports each method's excess error above this human--human reference. Relative to Zero-shot, \method{} closes 86\% of the gap in reliability error, 81\% in validity error, and 83\% in construct-level JSD. It remains measurably above human resampling error, so this empirical reference should not be interpreted as an attainable universal lower bound.

\begin{table}[t]
    \caption{ESS11 M1 gap to human resampling (Qwen). Excess error is model error minus the mean human--human error over 1,000 stratified resample pairs; lower is better. Gap closed is $(E_{\mathrm{ZS}}-E_{\mathrm{FR}})/(E_{\mathrm{ZS}}-E_{\mathrm{human}})$; higher is better.}
    \label{tab:humanref}
    \centering\footnotesize
    \setlength{\tabcolsep}{3.5pt}
    \begin{tabular*}{\textwidth}{@{\extracolsep{\fill}}lrrrrrr@{}}
        \toprule
        & Human & \multicolumn{4}{c}{Excess error above human resampling} & FR gap \\
        \cmidrule(lr){3-6}
        Metric & error & Zero-shot & \singleft{} & \sequentialft{} & \method{} & closed \\
        \midrule
        $\alpha$ MAE & .007 & .500 & .356 & .077 & \textbf{.071} & 85.9\% \\
        AVE MAE & .007 & .342 & .265 & .068 & \textbf{.065} & 81.1\% \\
        Item JSD & .007 & .067 & \textbf{.017} & .037 & \textbf{.017} & 74.4\% \\
        Construct JSD & $<$.001 & .124 & .037 & .047 & \textbf{.021} & 82.9\% \\
        \bottomrule
    \end{tabular*}
\end{table}

\textbf{Component and sampling ablation.} Table~\ref{tab:anchorablation} follows the path from Zero-shot and \singleft{} to a complete-response proposal and its \mcjp{}-corrected outputs. The proposal supplies cross-item dependence; projection brings its item margins closer to those learned by \singleft{}. At 16 candidates, this lowers construct JSD from .028 to .022. Increasing the projected candidate pool from one to 16 improves $\alpha$ and AVE errors, although not every metric changes monotonically. Because \mcjp{} needs complete-response proposals, these rows compare components and sampling budgets rather than successive fine-tuning checkpoints. Appendix~\ref{app:mechanism} gives the full candidate-budget and anchor-path audits.

\begin{table}[t]
    \caption{ESS11 M1 component and candidate-sampling ablation (Qwen; lower is better). The joint proposal uses 16 samples without projection; \mcjp{} uses learned Stage-1 margins. Rows with different candidate counts are frozen-model re-evaluations.}
    \label{tab:anchorablation}
    \centering\footnotesize
    \begin{tabular*}{\textwidth}{@{\extracolsep{\fill}}lrrrr@{}}
        \toprule
        Configuration & $\alpha$ MAE & AVE MAE & Item JSD & Construct JSD \\
        \midrule
        Zero-shot & .507 & .348 & .075 & .124 \\
        \singleft{} & .363 & .272 & \textbf{.024} & .037 \\
        Joint proposal (16 samples) & \textbf{.076} & \textbf{.070} & .027 & .028 \\
        \mcjp{} (1 sample) & .087 & .080 & \textbf{.024} & .022 \\
        \mcjp{} (16 samples) & .078 & .071 & \textbf{.024} & \textbf{.022} \\
        \bottomrule
    \end{tabular*}
\end{table}

\section{Related Work}
\label{sec:related}

\paragraph{Simulating survey responses.}
Prompted virtual samples established the appeal of LLM-based survey simulation \citep{argyle2023out}, while first-token specialization and subpopulation fine-tuning improved single-question response distributions \citep{cao2025specializing,suh2025scaled}. Background-shift alignment further targets how those distributions change across groups \citep{huang2026distribution}. Work on virtual respondents and restricted closed-ended generation broadens the task and clarifies the importance of response format \citep{zhao2026virtual,ahnert2026survey}. Our distinction is to train on linked questionnaires while separately preserving calibrated item rates, then evaluate both properties on held-out constructs and countries.

\paragraph{Joint structure and calibration.}
\citet{williams2026beyond} show that good item margins can conceal poor cross-item correlations; \citet{krsteski2026valid} study the complementary use of limited human data to rectify population estimates. We instead project a learned complete-response distribution toward margins predicted without held-out answers. The underlying I-projection and iterative fitting have a long statistical history \citep{csiszar1975idivergence,deming1940least}; our contribution is their use to couple two specialized survey LLMs while retaining whole-response candidates.

\section{Discussion}

The two learned models contribute different information: one-dimensional response rates and linked answer patterns. \mcjp{} combines them by changing first-order potentials while preserving the proposal's conditional odds ratios in the ideal projection. It never trains on $\alpha$, AVE, JSD, or downstream profit. Marginal validation, support adjustment, and effective sample size help diagnose when the projected answer pool cannot realize the intended rates; mixed gains with Ministral underline the importance of anchor quality.

Our evidence is limited to closed-ended batteries, two 8--9B backbones, one fine-tuning seed, and three generation rollouts. Finite candidate support weakens the ideal full-support KL guarantee (Appendix~\ref{app:finite}); country holdout does not establish transfer to every demographic intersection. The method also requires two fine-tunes and multiple proposal samples. Finally, $\alpha$ and AVE are descriptive diagnostics, not proof that synthetic respondents can replace human validation or support causal and consequential population claims.

\section{Conclusion}

Survey simulation should be evaluated at the level at which surveys are observed: linked answers from respondents. \method{} uses two ordinary LoRA adapters and one constrained distributional operation: the marginal model estimates how often answers occur, the respondent model estimates which answers occur together, and \mcjp{} finds the closest proposal distribution consistent with the calibrated rates. Across two surveys, two backbones, and three transfer regimes, this separation improves multi-item structure without giving up single-item accuracy. The remaining gap to human resampling shows that virtual respondents are not solved; the contribution is a formulation in which both improvement and failure are measurable.

\clearpage
\subsection*{AI use statement}
Generative AI assisted with research planning and methodology, experiment code, mathematical claims and proof drafts, literature search, figures, result interpretation, and manuscript drafting and editing. The authors checked source papers, code, equations, and numerical claims against saved artifacts and take responsibility for all content; no AI system is an author.

\subsection*{Ethics statement}
We use public-use ESS/TALIS files and de-identified responses from our Little-Treat survey under their respective data-use terms. Synthetic responses may distort minority views or be mistaken for real population evidence; they must not replace human surveys in consequential decisions. We will not release non-redistributable respondent records.

\subsection*{Reproducibility statement}
Appendix~\ref{app:repro} documents splits, prompts, training, sampling, and metrics; Appendix~\ref{app:theory} gives proofs. We will release code and aggregate outputs, while raw survey files remain with their providers.

\bibliography{references}
\bibliographystyle{iclr2027_conference}

\appendix
\section{Complete Theory}
\label{app:theory}

This appendix distinguishes the ideal population version of \mcjp{} from the finite-candidate implementation. The first yields exact KL identities; the second adds support approximation and Monte Carlo error.

\subsection{Setup and assumptions}

Fix a populated projection cell $g$ and omit it from notation. A complete questionnaire is $Y=(Y_1,\ldots,Y_m)\in\Omega=\prod_j\{1,\ldots,K_j\}$. Let $T$ be the cell-level human joint distribution, $P$ the cell-level mixture proposal in Eq.~\ref{eq:mixtureproposal}, $a$ the Stage-1 cell-marginal vector, and $t$ the true marginal vector. Remove the final redundant category for each item and define
\begin{equation}
 \varphi(Y)=\left(\mathbb 1\{Y_j=k\}\right)_{j=1:m,k=1:K_j-1}\in\mathbb R^d,
 \qquad t=\E_T\varphi(Y).
\end{equation}
For a feasible mean parameter $u$, let $\mathcal C(u)=\{Q:\E_Q\varphi=u\}$ and
\begin{equation}
 Q_u=\arg\min_{Q\in\mathcal C(u)}\kl(Q\|P).
 \label{eq:app_proj}
\end{equation}
We use the following standard conditions.
\begin{enumerate}[leftmargin=*]
 \item[(A1)] $T\ll P$: every human response pattern has positive proposal probability.
 \item[(A2)] $a$ and $t$ lie in the relative interior of the marginal polytope induced by the support of $P$.
 \item[(A3)] For the quantitative error bound, $\nabla^2 A(\lambda)=\operatorname{Cov}_{Q_\lambda}[\varphi(Y)]\succeq\mu I$ along the segment joining the dual parameters of $a$ and $t$, for some $\mu>0$.
\end{enumerate}
Dropping one category per item is necessary for A3 because a full one-hot vector is linearly dependent.

\subsection{From respondent training to a cell-level proposal}

The respondent objective is optimized for individual backgrounds and randomized item orders, whereas \mcjp{} acts on their population mixture.  This aggregation does not create an unaccounted risk term.  Let $w_i$ be normalized weights inside cell $g$, let $T_i$ be the human conditional joint for background $x_i$, and let $P_i^\pi$ be the order-specific respondent proposal.  Define
\begin{equation}
 T_g=\sum_iw_iT_i,
 \qquad
 P_g=\sum_iw_i\E_{\pi\sim\Pi}P_i^\pi.
\end{equation}
Joint convexity of KL and Jensen's inequality give
\begin{equation}
 \kl(T_g\|P_g)
 \le \sum_iw_i\kl\!\left(T_i\middle\|\E_\pi P_i^\pi\right)
 \le \sum_iw_i\E_\pi\kl(T_i\|P_i^\pi).
 \label{eq:app_mixture_risk}
\end{equation}
The final quantity is the population analogue of the order-randomized autoregressive loss, up to the entropy of $T_i$.  Thus Stage 2 controls an upper bound on the respondent-model term used below; it does not require any one display order to be canonical.

\subsection{Unique exponential-tilt solution}

\paragraph{Theorem A.1.}
Under A1--A2, Eq.~\ref{eq:app_proj} has a unique solution of the form
\begin{equation}
 Q_u(y)=P(y)\exp\{\lambda_u^\top\varphi(y)-A(\lambda_u)\},
 \quad
 A(\lambda)=\log\E_P\exp\{\lambda^\top\varphi(Y)\},
 \label{eq:app_tilt}
\end{equation}
where $\nabla A(\lambda_u)=u$.

\paragraph{Proof.}
Introduce Lagrange multipliers for normalization and the mean constraint in Eq.~\ref{eq:app_proj}.  Differentiation with respect to each positive $Q(y)$ gives
\begin{equation}
 \log\frac{Q(y)}{P(y)}=\lambda_u^\top\varphi(y)-c.
\end{equation}
Normalization yields $c=A(\lambda_u)$, and differentiating the log partition function gives $\nabla A(\lambda_u)=\E_{Q_u}\varphi=u$.  KL is strictly convex in $Q$ on the support of $P$, while $\mathcal C(u)$ is convex; the feasible stationary point is therefore unique. \hfill$\square$

\subsection{Exact decomposition and non-degradation}
\label{app:decomposition}

\paragraph{Theorem A.2.}
Under A1--A2,
\begin{align}
 \kl(T\|Q_a)&=\kl(T\|Q_t)+\kl(Q_t\|Q_a), \label{eq:app_decomp1}\\
 \kl(T\|P)&=\kl(T\|Q_t)+\kl(Q_t\|P). \label{eq:app_decomp2}
\end{align}

\paragraph{Proof.}
Equation~\ref{eq:app_tilt} implies
\begin{equation}
 \log\frac{Q_t(y)}{Q_a(y)}
 = (\lambda_t-\lambda_a)^\top\varphi(y)-A(\lambda_t)+A(\lambda_a),
\end{equation}
an affine function of $\varphi(y)$.  Because $T$ and $Q_t$ share the same true mean parameter $t$,
\begin{equation}
 \E_T\log\frac{Q_t(Y)}{Q_a(Y)}
 =\E_{Q_t}\log\frac{Q_t(Y)}{Q_a(Y)}
 =\kl(Q_t\|Q_a).
\end{equation}
Adding $\E_T\log(T/Q_t)$ proves Eq.~\ref{eq:app_decomp1}.  Likewise, $\log(Q_t/P)$ is affine in $\varphi$ and $T,Q_t\in\mathcal C(t)$, so
\begin{equation}
 \E_T\log\frac{Q_t(Y)}{P(Y)}
 =\E_{Q_t}\log\frac{Q_t(Y)}{P(Y)}=\kl(Q_t\|P),
\end{equation}
which proves Eq.~\ref{eq:app_decomp2}. \hfill$\square$

Combining the identities gives
\begin{equation}
 \kl(T\|Q_a)=\kl(T\|P)-\kl(Q_t\|P)+\kl(Q_t\|Q_a).
 \label{eq:app_gaincondition}
\end{equation}
If $a=t$, then $Q_a=Q_t$ and
\begin{equation}
 \kl(T\|Q_a)=\kl(T\|P)-\kl(Q_a\|P)\le\kl(T\|P).
\end{equation}
For imperfect margins, projection improves on the respondent model if and only if
\begin{equation}
 \kl(Q_t\|Q_a)<\kl(Q_t\|P).
\end{equation}
This condition is important when interpreting the Ministral experiments: a weak Stage-1 anchor can cost more than the correction gains.

\subsection{Component-to-fusion error bound}

Let $A^*$ be the convex conjugate of $A$.  Exponential-family duality gives
\begin{equation}
 \kl(Q_t\|Q_a)=D_{A^*}(t\|a).
\end{equation}
Under A3, $A^*$ is $1/\mu$-smooth, hence
\begin{equation}
 \kl(Q_t\|Q_a)\le\frac{1}{2\mu}\|t-a\|_2^2.
\end{equation}
Together with Theorem~A.2,
\begin{equation}
 \kl(T\|Q_a)\le R_{\mathrm{dep}}+\frac{1}{2\mu}\|t-a\|_2^2,
 \qquad R_{\mathrm{dep}}=\kl(T\|Q_t).
 \label{eq:app_meanbound}
\end{equation}
For each item, Pinsker's inequality gives $\|t_j-a_j\|_2^2\le\|t_j-a_j\|_1^2\le2\kl(t_j\|a_j)$.  Therefore
\begin{equation}
\boxed{\kl(T\|Q_a)\le R_{\mathrm{dep}}+
 \frac{1}{\mu}\sum_{j=1}^{m}\kl(t_j\|a_j).}
\label{eq:app_componentbound}
\end{equation}
Because $R_{\mathrm{dep}}=\kl(T\|P)-\kl(Q_t\|P)\le\kl(T\|P)$, we also obtain the directly trainable bound
\begin{equation}
 \boxed{\kl(T\|Q_a)\le \kl(T\|P)+
 \frac{1}{\mu}\sum_{j=1}^{m}\kl(t_j\|a_j).}
 \label{eq:app_trainablebound}
\end{equation}
The second term is the population risk estimated by Stage-1 cross-entropy, while Eq.~\ref{eq:app_mixture_risk} connects the first to Stage 2.  If both component risks converge to zero and $\inf_n\mu_n>0$, then the fused joint distribution is forward-KL consistent.

\subsection{Exact recovery and interaction preservation}

\paragraph{Theorem A.3 (exact recovery).}
Suppose the respondent model differs from the human distribution only through univariate potentials:
\begin{equation}
 \log\frac{T(y)}{P(y)}=c+\sum_{j=1}^{m}u_j(y_j).
 \label{eq:app_unipotential}
\end{equation}
If $a=t$, then $Q_a=T$.

\paragraph{Proof.}
Equation~\ref{eq:app_unipotential} places $T$ in the exponential family generated from base measure $P$ and sufficient statistic $\varphi$.  It also belongs to $\mathcal C(t)$.  By uniqueness in Theorem~A.1, it is the unique exponential tilt with those margins, so $T=Q_t=Q_a$. \hfill$\square$

\paragraph{Theorem A.4 (conditional odds-ratio invariance).}
For two items $r,s$, options $u,u',v,v'$, and fixed responses $z$ to all remaining items, define
\begin{equation}
 \operatorname{OR}^{Q}_{rs}(u,u';v,v'\mid z)=
 \frac{Q(u,v,z)Q(u',v',z)}{Q(u,v',z)Q(u',v,z)}.
\end{equation}
Whenever all probabilities are positive,
\begin{equation}
 \operatorname{OR}^{Q_a}_{rs}(u,u';v,v'\mid z)=
 \operatorname{OR}^{P}_{rs}(u,u';v,v'\mid z).
\end{equation}

\paragraph{Proof.}
The ratio $Q_a(y)/P(y)$ in Eq.~\ref{eq:app_tilt} factorizes into one term for each item.  Every factor appears once in the numerator and once in the denominator of the cross-product ratio and cancels. \hfill$\square$

Thus the ideal projection changes only first-order log-linear potentials.  It need not preserve Pearson correlations, whose values depend on both margins and interactions.

\subsection{What is and is not unbiased}

Let the Stage-1 anchor $\widehat a$ be random over training samples and suppose exact feasible projection ensures $\E_{Q_{\widehat a}}\varphi=\widehat a$.

\paragraph{Theorem A.5 (marginal and additive-estimand unbiasedness).}
If $\E_{\mathrm{train}}\widehat a=t$, then
\begin{equation}
 \E_{\mathrm{train}}\E_{Q_{\widehat a}}\varphi=t.
\end{equation}
For every additive survey statistic $f(Y)=c+\sum_j f_j(Y_j)$,
\begin{equation}
 \E_{\mathrm{train}}\E_{Q_{\widehat a}}f(Y)=\E_T f(Y).
\end{equation}

\paragraph{Proof.}
The first statement follows by iterated expectation.  Every function of a finite categorical item is a linear combination of its one-hot indicators; linearity proves the second. \hfill$\square$

This covers item means, response proportions, total-score means, and construct-average means.  It does \emph{not} imply $\E[Q_{\widehat a}]=Q_t$, because projection is nonlinear in $\widehat a$.  It therefore does not establish unbiasedness of correlations, Alpha, AVE, or complete joint cell probabilities.

Given fixed candidate weights $w_{1:M}$, randomized systematic resampling is conditionally unbiased: if $N_i$ is the number of copies of candidate $i$ in $N$ outputs, then $\E[N_i\mid w]=Nw_i$.  Hence sample averages of any $g(Y)$ equal the weighted candidate expectation in expectation, although they retain Monte Carlo variance.

\subsection{Finite candidate support}
\label{app:finite}

The implementation uses an empirical support $\mathcal S_M$ consisting of complete respondent-model candidates. Let $\delta_{\mathrm{miss}}=T(\mathcal S_M^c)$, $T_M=T(\cdot\mid\mathcal S_M)$, and $t_M=\E_{T_M}\varphi$. Let $\widetilde a$ be the feasible support-adjusted anchor, $Q^M_{\widetilde a}$ the empirical \mcjp{} solution, and $Q^M_{t_M}$ the corresponding oracle. Under the empirical-support analogue of A3 with constant $\mu_M$,
\begin{equation}
 \kl(T_M\|Q^M_{\widetilde a})\le
 R_{\mathrm{dep},M}+\frac{\|\widetilde a-t_M\|_2^2}{2\mu_M},
\end{equation}
where $R_{\mathrm{dep},M}=\kl(T_M\|Q^M_{t_M})$.  Since $Q^M_{\widetilde a}$ has zero probability outside $\mathcal S_M$, full-population forward KL may be infinite.  Total variation remains bounded:
\begin{equation}
 \operatorname{TV}(T,Q^M_{\widetilde a})\le\delta_{\mathrm{miss}}+
 \sqrt{\frac12\left(R_{\mathrm{dep},M}+
 \frac{\|\widetilde a-t_M\|_2^2}{2\mu_M}\right)}.
 \label{eq:app_tv}
\end{equation}
Writing $\delta_{\mathrm{support}}=\|\widetilde a-\widehat a\|_2$ and $\varepsilon_{\mathrm{stage1}}=\|\widehat a-t\|_2$ gives
\begin{equation}
 \|\widetilde a-t_M\|_2\le
 \delta_{\mathrm{support}}+\varepsilon_{\mathrm{stage1}}+
 \sqrt d\,\delta_{\mathrm{miss}}.
\end{equation}
Suppose the $N_g$ candidate questionnaires are conditionally independent draws (not necessarily identically distributed) and every required item--option event has probability at least $p_{\min}>0$ in every draw.  A union bound yields
\begin{equation}
 \Pr(\text{any required event absent})\le d'\exp(-N_gp_{\min}),
\end{equation}
where $d'$ is the number of required positive-margin events.  This guarantees only marginal feasibility, not coverage of all exponentially many questionnaires.

\subsection{Why item marginals cannot identify a respondent}

For two binary items, define $T(0,0)=T(1,1)=1/2$ and $R(0,1)=R(1,0)=1/2$.  Every item marginal agrees, but the supports are disjoint and $\jsd(T,R)=\log2$.  Conversely, the data-processing inequality for $f$-divergences implies
\begin{equation}
 \jsd(g_{\#}T,g_{\#}Q)\le\jsd(T,Q)
\end{equation}
for every item subset or construct mapping $g$.  Complete-joint closeness implies low-dimensional closeness; low-dimensional closeness does not imply complete-joint closeness.

\subsection{Psychometric summaries}

Cronbach's Alpha is a nonlinear covariance functional.  Scale $r$ items to $[0,1]$ and let $\Sigma_R$ be their covariance under distribution $R$:
\begin{equation}
 \alpha(R)=\frac{r}{r-1}\left(1-
 \frac{\operatorname{tr}(\Sigma_R)}{\mathbf 1^\top\Sigma_R\mathbf 1}\right).
\end{equation}
On finite response spaces, total variation controls all bounded first and second moments.  Consequently, Alpha is locally continuous whenever total-score variance is bounded away from zero.  This is a regularity statement, not a uniform guarantee: near a degenerate total score, the ratio can be unstable.  AVE additionally depends on the stability of a fitted factor model; local continuity requires item variances bounded away from zero, a unique CFA solution, and a nonsingular Jacobian of its first-order conditions.  We therefore make no unconditional Alpha or AVE claim.  They are held-out structural diagnostics that are neither constrained nor optimized by \mcjp{}.

\section{Prompts and Training Details}
\label{app:implementation}

\subsection{Construct-free prompt protocol}

\begin{figure}[t]
    \centering
    \includegraphics[width=0.96\textwidth]{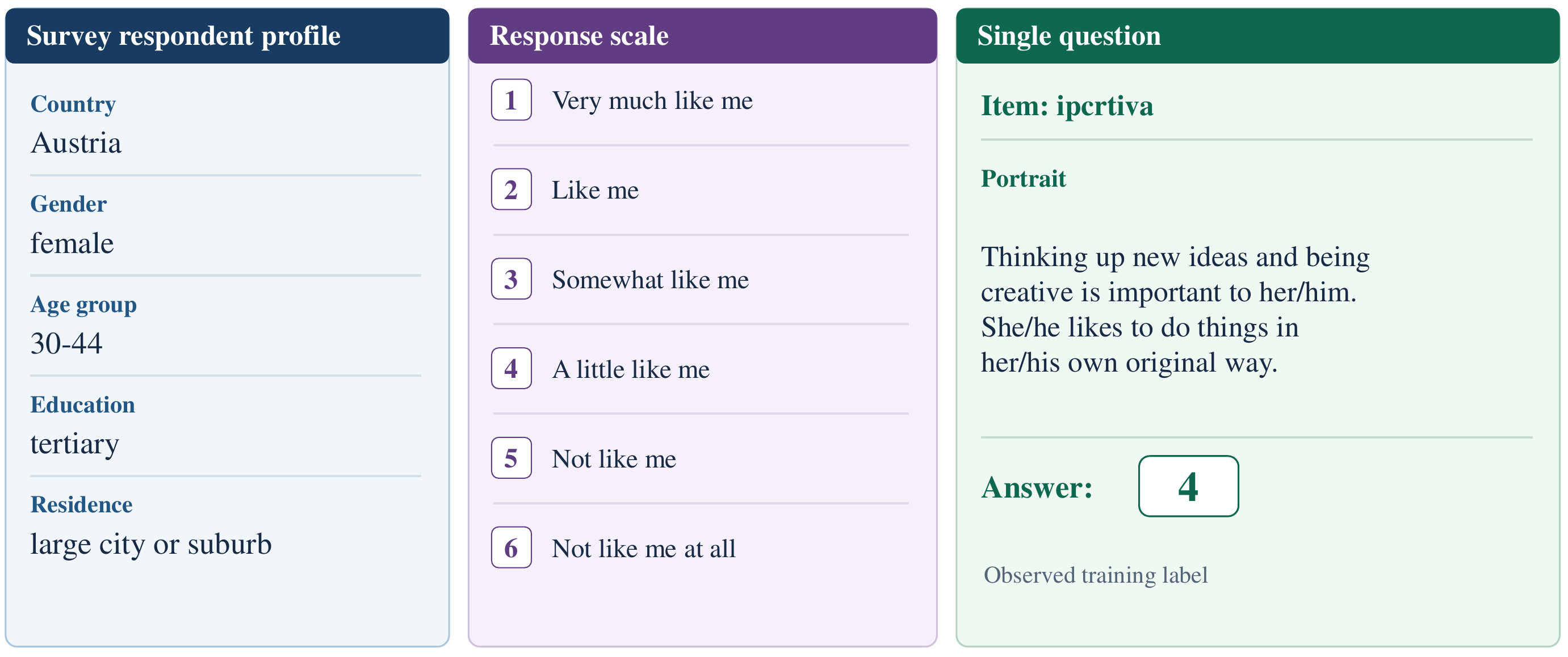}
    \caption{Stage-1 single-item training prompt, illustrated with one ESS11 response. The model sees the respondent profile, option scale, and one portrait; cross-entropy supervises the observed answer token. No previous question or construct name is supplied.}
    \label{fig:prompt_single}
\end{figure}

\begin{figure}[t]
    \centering
    \includegraphics[width=0.96\textwidth]{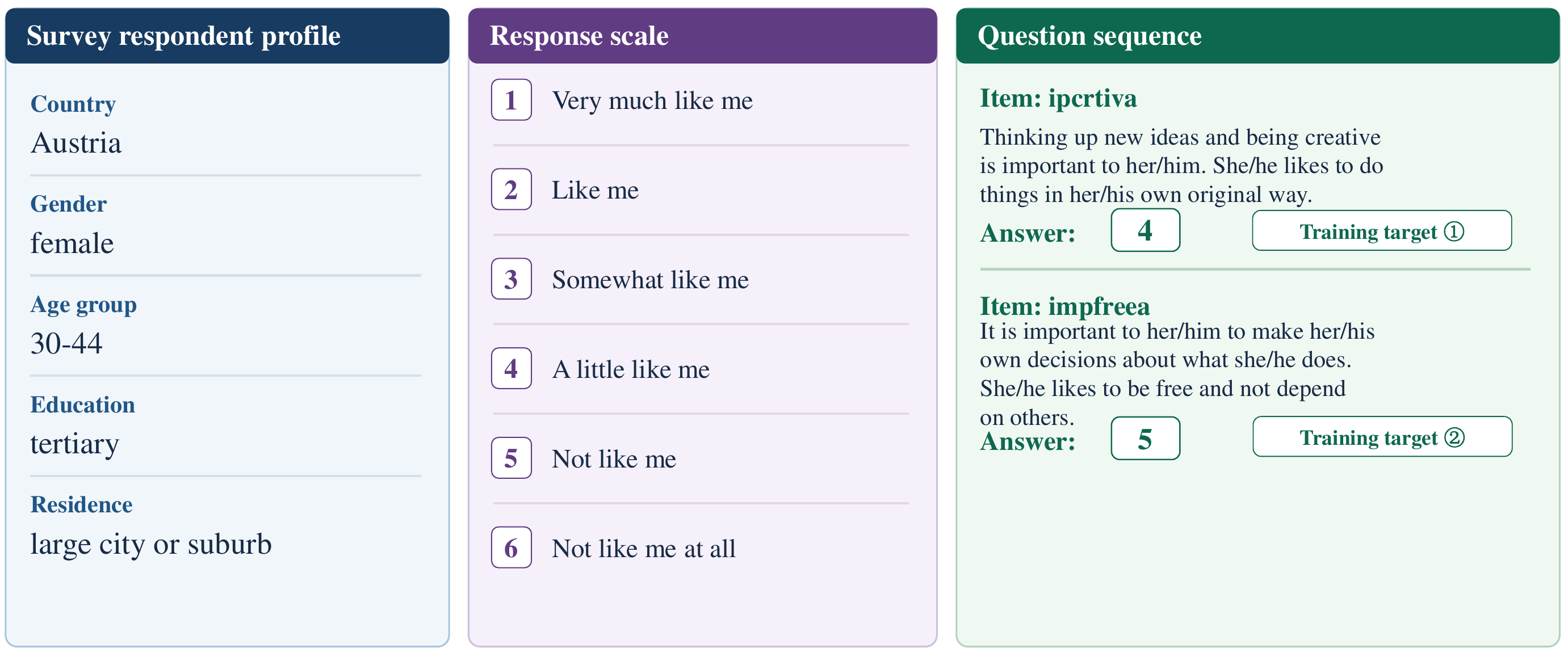}
    \caption{Stage-2 linked-question training prompt for the same respondent. Each preceding observed answer is available when the next answer is predicted, and losses at all answer positions are included in one respondent-level training example. Actual training randomizes item order and uses all eligible items, rather than only the two pictured here.}
    \label{fig:prompt_joint}
\end{figure}

Both stages use the same item wording and response scale.  Construct names and definitions are omitted.  The ESS11 template is schematically:
\begin{quote}\small
\texttt{Survey respondent profile}\\
\texttt{Country: [country]}\\
\texttt{Gender: [gender]}\\
\texttt{Age band: [age]}\\
\texttt{Education band: [education]}\\
\texttt{Domicile: [domicile]}\\[1mm]
\texttt{Response scale: 1 = Very much like me;}\\
\texttt{...; 6 = Not like me at all.}\\
\texttt{Answer as the same respondent throughout the questionnaire.}\\
\texttt{[Optional previous question and sampled/true answer history]}\\
\texttt{Question: [official item text]}\\
\texttt{Answer:}
\end{quote}
Stage 1 omits history (Figure~\ref{fig:prompt_single}). Stage 2 teacher-forces true history during training and uses generated history at inference (Figure~\ref{fig:prompt_joint}). The visuals show two ESS items and illustrative observed labels; they abbreviate the repeated instruction line in the serialized header above. TALIS replaces the background fields and uses its official four-option scale. Direct option-token scoring prevents prose, refusal, or formatting variation from being interpreted as a response.

\subsection{Optimization}

Both stages use LoRA rank 8, scaling 16, dropout 0.05, bfloat16 arithmetic, SDPA attention, gradient accumulation 16, and seed 20260727.  ESS uses learning rate $10^{-4}$, Stage-1 batch size 2 for one epoch, and Stage-2 batch size 1 for at most four epochs with patience 2.  TALIS uses batch size 1 for at most ten epochs with patience 2; the Qwen and Ministral learning rates are $2\times10^{-4}$ and $10^{-4}$, respectively.  Stage 2 averages validation loss over three sampled item orders.  Checkpoint selection uses development-country validation, never test-country labels.  Stage 1 selects item negative log-likelihood; Stage 2 selects respondent-level autoregressive negative log-likelihood.

For the final projection, the respondent model generates 16 candidate questionnaires per target respondent and IPF operates within population cells. Three outputs per target are drawn by randomized systematic resampling. All baselines also produce three final rollouts. ESS analysis weights and TALIS teacher weights are retained when aggregating human and generated distributions.

\clearpage

\section{Data Protocol Details}
\label{app:repro}

\subsection{ESS11}

Eligibility requires valid 1--6 responses for all 21 PVQ-21 items and known gender and age, leaving 42,551 respondents. Education and domicile nonresponse are explicit ``unknown'' prompt values. The train countries are disjoint from five development and five test countries. The training constructs are Universalism (3 items), Security (2), Achievement (2), and Hedonism (2). The unseen constructs are Self-Direction, Power, Stimulation, Conformity, Tradition, and Benevolence, each with two items. M1 uses 144 populated country--gender--age cells; M2 and M3 use 40 held-out-country cells.

\subsection{TALIS 2018}

The international teacher file contains 261,426 records, of which 245,450 have valid answers to all 12 self-efficacy items.  Classroom Management and Instruction contain four items each and are observed in training; Student Engagement contains four held-out items.  Thirty-three systems form the training pool, seven are development-only, and seven are test-only.  The frozen training sample has 12,000 teachers; each mode has 8,000 generated targets.  Distribution metrics group by country/system and gender while the prompt additionally contains education and first-career-choice status.

\section{Metric Definitions}
\label{app:metrics}

\paragraph{Item JSD.}
For each populated evaluation cell $g$ and item $j$, human and generated weighted histograms over response categories are normalized to distributions $p_{gj}$ and $q_{gj}$.  We compute
\begin{equation}
 \jsd(p,q)=\tfrac12\kl(p\|\tfrac12(p+q))+
 \tfrac12\kl(q\|\tfrac12(p+q)),
\end{equation}
then average over cells and items.  Empty cells are not invented; the frozen protocol uses only populated human cells.  A Dirichlet pseudocount of 0.5 is applied to the human reference histogram where specified by the saved protocol.

\paragraph{Construct JSD.}
For construct $c$, each respondent contributes one categorical tuple in the Cartesian product of its item response spaces.  We compare human and generated tuple histograms and average the resulting JSD after calculation across constructs.  Paired-construct JSD applies the same operation to all item tuples in $c\cup c'$.  It is not the JSD between an average item histogram and therefore cannot cancel item-specific errors.

\paragraph{Reliability and convergent validity.}
Cronbach's Alpha is computed from the respondent-level item covariance matrix for every official construct.  AVE is computed from standardized one-factor CFA loadings using $\mathrm{AVE}=r^{-1}\sum_{j=1}^{r}\lambda_j^2$.  We report the mean absolute difference between generated and human construct values, not the generated value alone.  Failed or degenerate fits are recorded rather than silently replaced.

\paragraph{Human sampling reference.}
Complete respondents, not individual answers, are resampled within country--gender--age strata with survey weights.  We run 1,000 replicates at the model evaluation sample size.  ``Human versus human'' compares two independent resamples and is an empirical sampling reference, not a universal information-theoretic lower bound.

\section{Two-Construct Results}

The main tables omit two-construct JSD for readability.  Table~\ref{tab:pairjsd} reports it separately.  TALIS evaluates only one construct in M1 and M3, so the metric is undefined in those modes.

\begin{table}[t]
\caption{Two-construct JSD. Lower is better. Bold is best and underline is second best within each dataset, backbone, and applicable mode.}
\label{tab:pairjsd}
\centering\footnotesize
\setlength{\tabcolsep}{4.0pt}
\renewcommand{\arraystretch}{1.08}
\begin{tabular*}{\textwidth}{@{\extracolsep{\fill}}llrrr@{}}
\toprule
Data / backbone & Method & M1 & M2 & M3 \\
\midrule
\multirow{4}{*}{ESS11 / Qwen}
& Zero-shot & .247 & .366 & .296 \\
& \mbox{Single FT} & \underline{.099} & \underline{.142} & \underline{.160} \\
& \mbox{Sequential FT} & .110 & .165 & .184 \\
& \method & \textbf{.059} & \textbf{.081} & \textbf{.123} \\
\midrule
\multirow{4}{*}{ESS11 / Ministral}
& Zero-shot & .291 & .376 & .340 \\
& \mbox{Single FT} & \underline{.119} & \underline{.150} & \underline{.182} \\
& \mbox{Sequential FT} & .179 & .171 & .230 \\
& \method & \textbf{.080} & \textbf{.086} & \textbf{.145} \\
\midrule
\multirow{4}{*}{TALIS / Qwen}
& Zero-shot & -- & .571 & -- \\
& \mbox{Single FT} & -- & .347 & -- \\
& \mbox{Sequential FT} & -- & \underline{.150} & -- \\
& \method & -- & \textbf{.115} & -- \\
\midrule
\multirow{4}{*}{TALIS / Ministral}
& Zero-shot & -- & .674 & -- \\
& \mbox{Single FT} & -- & .359 & -- \\
& \mbox{Sequential FT} & -- & \underline{.196} & -- \\
& \method & -- & \textbf{.123} & -- \\
\bottomrule
\end{tabular*}
\end{table}

\section{Additional Analyses}
\label{app:additional}

\subsection{Proposal budget}
\label{app:mechanism}

\begin{table}[t]
\caption{ESS11 M1 candidate-support audit.  Item JSD remains anchored while joint support and feasibility improve.}
\centering\footnotesize
\renewcommand{\arraystretch}{1.08}
\begin{tabular*}{\textwidth}{@{\extracolsep{\fill}}rrrrrr@{}}
\toprule
\shortstack[c]{Candidate\\questionnaires} & Item JSD & \shortstack[c]{Two-construct\\JSD} & \shortstack[c]{Maximum\\marginal error} & \shortstack[c]{Adjusted\\cells} & \shortstack[c]{Effective sample\\fraction} \\
\midrule
1 & .0244 & .0693 & .1952 & 469 & .330 \\
2 & .0244 & .0621 & .0748 & 201 & .422 \\
4 & .0244 & .0596 & .0190 & 76 & .506 \\
8 & .0244 & .0596 & .0112 & 18 & .555 \\
16 & .0244 & .0592 & .0013 & 2 & .578 \\
\bottomrule
\end{tabular*}
\end{table}

\subsection{Marginal-anchor interpolation}

Holding the same 16-candidate proposal fixed, we interpolate between its own margins ($\rho=0$) and the learned Stage-1 anchor ($\rho=1$).  The monotone improvement in the two joint metrics shows that the gain is produced by constrained fusion rather than by regenerating candidates.  Effective-sample fraction decreases because stronger correction necessarily concentrates weights.

\begin{table}[t]
\caption{ESS11 M1 anchor path.  The proposal and candidate support are fixed; only the target margin changes.}
\label{tab:anchorpath}
\centering\footnotesize
\renewcommand{\arraystretch}{1.08}
\begin{tabular*}{\textwidth}{@{\extracolsep{\fill}}rrrrrr@{}}
\toprule
$\rho$ & Item JSD & Construct JSD & Two-construct JSD & \shortstack[c]{Support\\adjustment} & \shortstack[c]{Effective sample\\fraction} \\
\midrule
0.00 & .0267 & .0280 & .0708 & .0000 & 1.000 \\
0.25 & .0251 & .0254 & .0661 & .0013 & .962 \\
0.50 & .0242 & .0240 & .0636 & .0027 & .860 \\
0.75 & .0240 & .0224 & .0606 & .0040 & .722 \\
1.00 & .0244 & .0217 & .0596 & .0053 & .578 \\
\bottomrule
\end{tabular*}
\end{table}

An analysis-only oracle-smoothed anchor derived from held-out labels reaches Item/construct/two-construct JSD of .0024/.0022/.0202.  Shuffling learned anchors across background cells instead yields .0450/.0226/.0622.  The oracle is not a deployable method; together, these controls isolate anchor quality and correct background assignment as the relevant mechanism.

\subsection{Controllability and response behavior}
\label{app:backgroundshift}

The ESS11 country-swap analysis pairs each source country with a target country and compares item-mean changes generated by changing only the prompt field with observed human differences. Across 216 country--item effects, Pearson correlation and cosine similarity are 0.655 and direction agreement is 0.724 for effects larger than 0.02. Removing all backgrounds worsens Alpha MAE from 0.078 to 0.102, AVE MAE from 0.071 to 0.094, Item JSD from 0.024 to 0.042, and pair JSD from 0.059 to 0.096.

We additionally evaluate targeted background changes on the 12 held-out ESS11 items. For each source--target rule $r$ and item $j$, $\Delta^{\mathrm{human}}_{rj}$ is the weighted difference between the corresponding real target and source groups. $\Delta^{\mathrm{model}}_{rj}$ is the change obtained by editing only the designated prompt fields for the same source respondents. Held-out human target responses are used only to score this audit. We report
\begin{equation}
    \operatorname{ShiftMAE}_c=
    \frac{1}{|\mathcal R_c|m}
    \sum_{r\in\mathcal R_c}\sum_{j=1}^{m}
    \left|\Delta^{\mathrm{model}}_{rj}-
    \Delta^{\mathrm{human}}_{rj}\right|,
\end{equation}
which jointly penalizes incorrect direction and magnitude in ESS response-scale points. Table~\ref{tab:shiftmae} compares the five contrasts shared by all three methods, with equal weight in the macro average.

A gender-only contrast was run for the two learned methods but excluded from the three-method aggregate because Zero-shot was not generated; \method{} and \singleft{} obtain .081 and .079, respectively. The differences in Table~\ref{tab:shiftmae} are not statistically resolved under one generation seed, and prompt-field edits do not identify causal demographic effects.

A respondent-bootstrap audit also compares extreme-response rate, central-response rate, straightlining, within-person standard deviation, entropy, item-correlation RMSE, and construct-profile correlations. The results are mixed across these secondary diagnostics: \method{} improves several profile-level errors over \singleft{} and \sequentialft{}, but not every response-style statistic. We therefore do not treat marginally constrained dependence as a complete behavioral model.

\subsection{Option-label robustness}

A 256-respondent ESS11 M1 pilot reverses the displayed option labels and measures how much each method's predicted distribution moves. Mean item-level shift JSD is 0.123 for \singleft{} and 0.112 for \method{}. Thus the complete method attenuates, but does not eliminate, sensitivity to option presentation. Because this is a small diagnostic rather than the frozen full-population protocol, we do not mix it into the primary tables.

\subsection{Synthetic-to-real downstream validity}

We also ask whether a complete synthetic questionnaire carries usable dependence for leave-one-item-out imputation. For each evaluation mode, an L2-regularized multinomial logistic model is trained on a 70\% synthetic split to predict each item from the other items and demographics, then tested on the matched 30\% held-out human split. In M1, mean NLL is 1.411 when trained on human questionnaires and 1.421 when trained on \method{} questionnaires. The small gap shows that the generated records retain useful multivariate signal, although this task-specific diagnostic is not evidence that synthetic data can replace human data.

The apparently high M2/M3 NLL of the human-trained reference is not a claim that synthetic data is superior to human data.  The reference classifier is trained on one finite human split under country shift, whereas synthetic generators pool inductive bias from model pretraining and the survey fine-tuning protocol.  The comparison is a task-specific validity diagnostic, not a replacement argument.

\subsection{Little-Treat decision simulation}
\label{app:littletreat}

We use our Little-Treat Consumption questionnaire. Of 488 complete records, 294 train the survey models, 97 are used for development, and 97 are held out for evaluation. We retain six attitude items, purchase frequency, and the most recent purchase-price range. The latter is treated as a proxy spending ceiling, not an elicited maximum willingness to pay. For a candidate price $p$, the joint synthetic answers estimate monthly demand per respondent $\widehat D_m(p)$ by weighting the frequency of respondents whose proxy ceiling is at least $p$. The same calculation on held-out human answers gives $D_{\mathrm{real}}(p)$.

The original saved pricing run used a unit cost of 8 CNY and stored each method's predicted profit curve. The pricing-and-stocking analysis in Figure~\ref{fig:commercial} is a subsequent recomputation from those curves, \emph{not} a new model-generation run or a preregistered cost comparison. For each price in \{10,20,40,80,150,300,600\} CNY, we recover predicted demand as the stored profit divided by $p-8$. At a hypothetical new unit cost $c$, each method selects $p_m=\arg\max_p(p-c)\widehat D_m(p)$ and stocks $Q_m=\widehat D_m(p_m)$ per respondent. For zero salvage, realized proxy profit per respondent is $p_m\min(Q_m,D_{\mathrm{real}}(p_m))-cQ_m$; unsold stock is $(Q_m-D_{\mathrm{real}}(p_m))_+$. Main-text totals multiply these per-respondent quantities by 1,000 for readability; only 97 respondents were actually held out. Figure~\ref{fig:commercial} converts costs and profits to USD equivalents at a fixed 7 CNY per USD solely for presentation; all decisions were computed in the original currency.

At costs of 50 and 150 CNY, \method{} has the highest profit point estimate, and its lead also remains under salvage values of 25\% or 50\% of unit cost. Zero-shot selects the same 300 CNY price but predicts 1.197 monthly purchases per respondent there, versus 0.317 in the held-out records, causing substantial overstock. In 10,000 paired bootstrap resamples of the 97 human test respondents with model decisions fixed, the 95\% intervals for \method{} minus \sequentialft{} include zero at both costs. These results illustrate sensitivity to the joint frequency--spending distribution; they do not establish a significant advantage over that baseline, causal price elasticity, or revenue realized in a live market.

\end{document}